\pdfoutput=1

\documentclass[11pt, a4paper]{article}

\usepackage{ACL2023}

\usepackage{times}
\usepackage{latexsym}
\usepackage{booktabs}
\usepackage[T1]{fontenc}
\usepackage{graphicx}
\usepackage{algorithm}
\usepackage{algpseudocode}
\usepackage{amsmath}

\usepackage[utf8]{inputenc}

\usepackage{microtype}

\usepackage{inconsolata}

\usepackage{array}

\newcolumntype{P}[1]{>{\centering\arraybackslash}p{#1}}

\begin{document}
\title{RoundTable: Leveraging Dynamic Schema and Contextual Autocomplete for Enhanced Query Precision in Tabular Question Answering}

\author{Pratyush Kumar$^{1}$\footnotemark[1] \,, Kuber Vijaykumar Bellad$^{1}$\footnotemark[1] \,, Bharat Vadlamudi$^{1}$\footnotemark[1] \,, Aman Chadha$^{2,3}$\footnotemark[3] \\
$^{1}$West Pharmaceutical Services, Inc. \\
$^{2}$Stanford University \\
$^{3}$Amazon GenAI \\
\small\texttt{\{pratyushk2011, kuber.bellad4, bharat0355\}@gmail.com; hi@aman.ai} \\}

\maketitle
\footnotetext[1]{\,Work does not relate to position at West.}
\footnotetext[3]{\,Work does not relate to position at Amazon.}

\begin{abstract}
With advancements in Large Language Models (LLMs), a major use case that has emerged is querying databases in plain English, translating user questions into executable database queries, which has improved significantly. However, real-world datasets often feature a vast array of attributes and complex values, complicating the LLM's task of accurately identifying relevant columns or values from natural language queries. Traditional methods cannot fully relay the dataset's size and complexity to the LLM. To address these challenges, we propose a novel framework that leverages Full-Text Search (FTS) on the input table. This approach not only enables precise detection of specific values and columns but also narrows the search space for language models, thereby enhancing query accuracy. Additionally, it supports a custom auto-complete feature that suggests queries based on the data in the table. This integration significantly refines the interaction between the user and complex datasets, offering a sophisticated solution to the limitations faced by current table querying capabilities.
This work is accompanied by an application \footnote{\url{https://github.com/pkscanvas/RoundTable}} for both Mac and Windows platforms, which readers can try out themselves on their own data.
\end{abstract}

\begin{figure*}[htp]
    \centering
    \includegraphics[width=\textwidth]{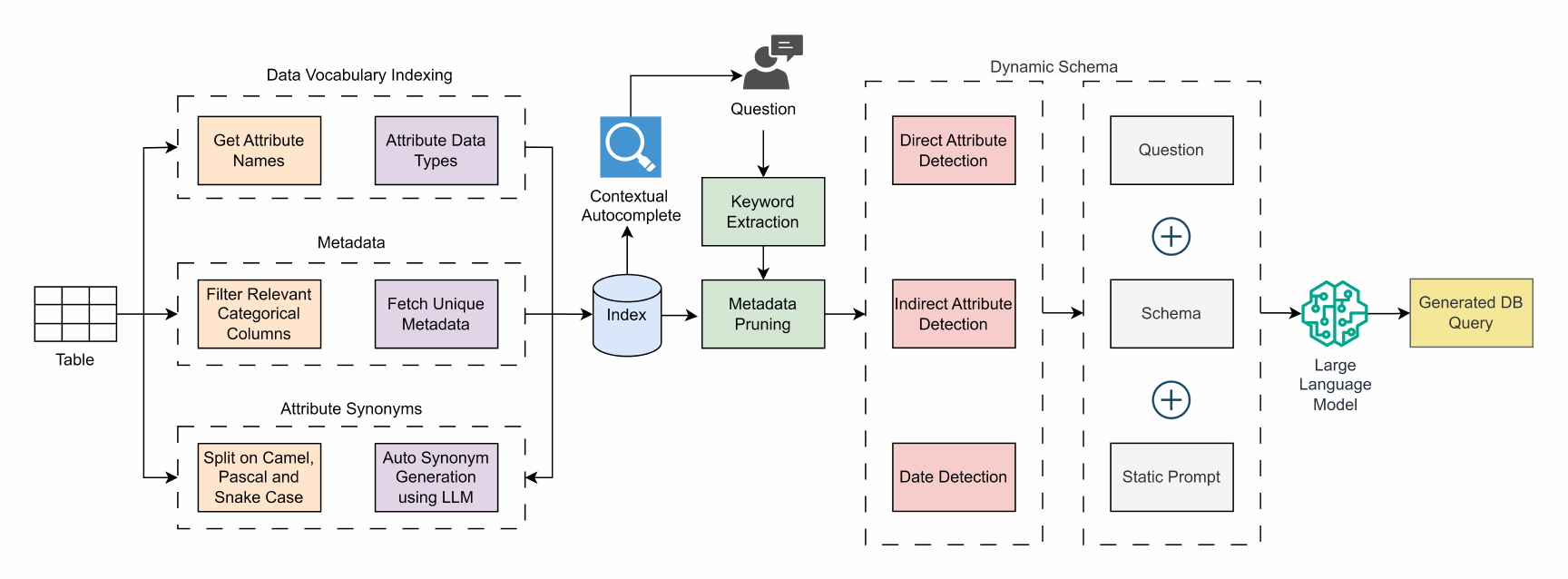}
    \caption{The architectural details are illustrated, depicting the three major components. 1. Data Vocabulary Indexing 2. Contextual Autocomplete 3. Dynamic Schema Generation }
    \label{fig:round table}
\end{figure*}

\section{Introduction}
The advent of Large Language Models (LLMs) has revolutionized the way we interact with data, offering unprecedented capabilities for processing and understanding natural language queries. As these models advance, their ability to translate complex user inquiries into executable database queries has significantly improved, making natural language processing an increasingly viable method for database interaction. This evolution holds the promise of democratizing data access, allowing users without much technical expertise in query languages to harness the power of vast datasets.

The trend is further cemented by OpenAI's recent addition of Advanced Data Analysis capabilities to ChatGPT. This innovation has the potential to democratize data access by making large databases more accessible to those who may not be as proficient in formal query languages.

The current landscape of TQA can be categorized into three broad approaches to the best of our knowledge:

\subsection{Model-Centric Approaches}

By training encoder-decoder models like TAPEX \citep{liu2022tapex_9} to deep learning-based models like MLM methods \citep{Herzig_2020_10}, the models can understand the subtleties of organized tabular data and provide answers based on it. These methods have a significant drawback in that the complete dataset needs to be loaded into memory, which makes them inappropriate for huge, complicated real-world datasets. Additionally, the context length restricts the quantity of data that may be supplied to the model by design, just like any other LLM.

\subsection{Reasoning and Tool-Enhanced Frameworks}

\citet{gemmell2023generate_13} shows how to incorporate tool synthesis techniques and generate responses using structured reasoning techniques like Chain of Thoughts prompting \citep{wei2023chainofthought_11}. These approaches break down the table into numerous phases and use structured reasoning over the table to determine the best potential answer to the user inquiry. In a similar vein, an LLM's ability to identify a potential remedy is enhanced using one or more tools. Similarly, for the model to complete the analysis, many of these approaches require the entire dataset to be provided. However, these methods have produced some excellent results on web-based tables like WikiSQL \citep{zhong2017seq2sql_12}, which are short in size and compatible with the majority of contemporary LLMs. However, it also renders these methods unfeasible for practical use in the real world.


\subsection{Natural Language to DB Query}

Converting natural language questions about the dataset into corresponding database queries by utilizing the general capabilities of the LLM, which include code generation in several programming languages \citep{cheng2023binding_4}, is another developing field of study. Target languages for tabular queries are often SQL and Pandas, which are among the most popular query languages and frameworks available today. This specific method of dataset querying has the benefit of being functional regardless of the dataset's location or size, allowing the generated query to be executed in a secure environment to obtain the answer and not being restricted to a fixed memory/token limit. This makes it useful for many large and complex real-world datasets.

Open-source LLMs such as Llama \citep{touvron2023llama}  and Mistral \citep{jiang2023mistral} have allowed businesses to control the data and model, creating new opportunities for TQA systems that prioritize security and privacy at enterprise level. Additionally, improvements in quantization methods such as GGUF\footnote{\url{https://github.com/ggerganov/ggml/blob/master/docs/gguf.md}}, GGML\footnote{\url{https://github.com/ggerganov/ggml}}, GPTQ \citep{frantar2023gptq_17}, and most significantly, frameworks like llama.cpp, have made it feasible to run these models on consumer-grade hardware, granting end users unrestricted, private, and cost-free access to these impressive models. Natural language processing is becoming a more practical approach for database interaction as these models develop and get better at translating complex user questions about a dataset into executable database queries.

This paper outlines a novel system that uses Full-Text Search (FTS) to address the difficulties associated with using natural language queries on huge datasets. Using an autocomplete tool, our method seeks to improve LLM accuracy, reduce compute burden, and make query generation easier. Through the establishment of a connection between structured databases and natural language comprehension, our system provides enhanced precision and efficiency when handling complex real-world data.


\section{Related Work}

\subsection{Enhancing Table Question Answering (TQA) through Advanced Frameworks and Techniques}

\textbf{ReAcTable and Chain of Table} both focus on improving the performance of TQA by introducing innovative frameworks that enhance the reasoning capabilities of LLMs with tables. 

The difficulty of Table Question Answering (TQA) at the nexus of data analytics and natural language processing is examined in this study. This assignment requires logical reasoning, a comprehension of data semantics, and analytical skills to answer queries in normal language based on tabular data. To handle TQA activities, the study presents ReAcTable, a framework that improves on the ReAct paradigm \citep{zhang2023reactable_1}. Despite complexity and processing needs, it offers a revolutionary framework for TQA problems that greatly improves performance using a combination of huge language models, external code executors, and creative majority voting procedures.

 \citet{wang2024chainoftable_2} investigate how to include LLMs into table-based reasoning, which is an essential component of tasks such as fact verification and table-based question answering. The Chain-Of-Table structure, which is introduced in this research, greatly improves LLMs' capacity to handle and analyze tabular data. Because one must comprehend the semantics of the queries as well as the semi-structured tabular data, table-based reasoning is difficult. Previous methods, such as Chain-of-Thought, enhanced the efficiency of LLMs by integrating textual reasoning chains, but they had trouble utilizing tabular data effectively. The reasoning chain's use of tabular data as a stand-in for intermediate concepts is suggested by the framework. It directs LLMs to produce operations and update the table iteratively to depict a complex reasoning chain. Through the formation of a chain that illustrates the thought process for a particular problem, this technique enables LLMs to dynamically design the subsequent operation based on past findings. Adding columns, choosing rows, grouping, and other atomic operations are frequently utilized in SQL and DataFrame development, and Chain-Of-Table makes ad-vantage of these operations. Table manipulation and thinking are made easier by these operations in different ways. The latest table state and prior operations are used to dynamically design activities in the framework. To update the table, it also consists of a stage for creating arguments for the scheduled table operations. Chain-Of-Table demonstrated state-of-the-art performance when tested on three benchmark datasets: WikiTQ, FeTaQA, and TabFact. The tests in the trials utilized few-shot demo samples for in-context learning and the GPT-3.5 and PaLM 2-S1 models for testing. With Chain-Of-Table, LLMs' reasoning powers in table-based tasks are greatly improved. It offers a new way to integrate table data directly into the reasoning chain, enabling dynamic, iterative processing, and operation planning.

\textbf{API-Assisted Code Generation for Question Answering on Varied Table Structures} presents a unified TQA architecture that adapts to different table structures, which can be seen as complementary to the efforts of enhancing TQA through more adaptable and flexible frameworks.

\citet{cao2023apiassisted_3} address the problem of adapting (TQA) to different table structures. These structures frequently call for logical forms specific to a certain domain. In comparison to earlier methods, the research presents a unified TQA architecture that delivers notable advantages. Although tables are frequently used to store and retrieve data, there is significant variation in their structure, which makes TQA tasks difficult. The goal of the suggested framework is to create a uniform method for handling TQA jobs with various table architectures while utilizing Python as an intermediary language inside the Python Pandas module. This involves combining differently structured tables into a single multi-index data frame representation. Using code generation models to prompt TQA questions to be translated into Python scripts, an executable intermediate language. Adding assistant API calls will allow the framework to do more than just Python Pandas. It will be able to query external knowledge and carry out a variety of other activities. Using CODEX and STARCODER code generation models, the framework was assessed on four TQA datasets: WikiTableQuestions (WikiTQ), HiTab, AIT-QA, and Spider. It outperformed cutting edge few-shot baselines on the HiTab, AIT, and Spider datasets with notable gains. Using a unifying multi-index format to describe differently structured tables and bespoke auxiliary API methods that are trained to be utilized by the LLM in context are two aspects of the methodology. It addresses the variety of table structures and expands on the capabilities of LLMs in this field by presenting an innovative method to TQA. The modular architecture of the framework, along with the use of Python and APIs, show how flexible and expandable it can be.

\subsection{Integrating Symbolic Languages and Operations with LLMs for TQA}

\textbf{Binding Language Models in Symbolic Languages and LLMs are Versatile Decomposers: Decompose Evidence and Questions for Table-based Reasoning} both explore the integration of symbolic languages (like SQL) and operations with LLMs to improve TQA. These studies focus on enhancing interpretability, resilience, and performance by bridging the gap between natural language processing and symbolic reasoning.

BINDER \citep{cheng2023binding_4}, is neural-symbolic and does not require training. This framework extends the capabilities of computer languages such as Python and SQL by integrating language models (LMs) with them to address a variety of problems. With a focus on table question answering, BINDER seeks to enhance the interpretability and resilience of end-to-end neural techniques in NLP tasks. GPT-3 Codex is the LM used by BINDER. Two phases make up its operation: parsing and execution. Codex creates a program from the input question during the parsing step. It does this by recognizing questions that the programming language cannot answer on its own and by creating the necessary API calls. Codex uses these API calls to carry out operations during the execution phase. In general, to improve the interpretability, resilience, and adaptability of NLP systems in tasks such as table question answering, the study offers a potential technique to integrate language models with symbolic programming languages. Its efficacy is, therefore, partially reliant on the underlying LM, and more research is needed to see whether it can be used generally to a range of NLP tasks.

\citet{ye2023large_5} investigates how to enhance table-based reasoning through the application of LLMs. The primary goal is to employ LLMs to break down the questions and the evidence (big tables) into more digestible chunks, which will help with better thinking and response creation. DATER (Decompose evidence and questions for efficient Table-based Reasoning) is a revolutionary approach introduced by the authors. Using this method, lengthy tables are broken down into pertinent sub-evidence and difficult questions into more manageable sub-questions. This deconstruction is accomplished by means of a "parsing-execution-filling" technique, intermediate SQL query generation, and in-context learning with LLMs. Overall, the work offers a novel method for improving table-based reasoning by utilizing LLMs' deconstruction powers. By addressing major issues with managing huge tables and complex queries, this approach greatly enhances interpretability and performance. Its use of LLMs and the difficulty of its implementation, however, are significant drawbacks. It is still to be determined whether this method can be used to different kinds of data and reasoning problems.

\subsection{Novel Methodologies and Tools for TableQA}

\textbf{TableQAKit} offers a comprehensive toolkit that integrates various techniques and LLMs for TableQA, serving as a practical resource for researchers and practitioners.

The toolkit by \citet{lei2023tableqakit_7} attempts to offer a uniform platform that integrates common techniques, such as using LLMs, with different TQA datasets. It blends various techniques and LLMs for these tasks and has a large variety of TQA datasets. Incorporates a fresh approach—that is, LLM-based methodologies for organized QA assignments. Provides a user-friendly interactive interface with visible operations and extensive data. The models, datasets, and source code are accessible to the public. A complete and cohesive toolbox that works with nearly any TQA situation. TableQAEval is a benchmark for LLM-based TQA that is the first multi-type long-context benchmark. Tools for data visualization that enable many table formats, multi-modal data, and easier data interaction. The discipline of NLP has benefited greatly from TableQAKit, especially in the domain of TQA. It incorporates LLMs in a new approach, gives a uniform platform to handle different TQA tasks, and offers helpful benchmarks to assess LLM capabilities in this situation. Because of its open-source nature and intuitive design, the toolkit is a useful and easily available resource for scholars and practitioners working in the subject.

\textbf{HRoT} presents a novel approach for improving Table-Text Hybrid QA by leveraging a hybrid prompt strategy, which can be categorized as a unique methodology aimed at a specific aspect of TQA involving hybrid data.

Advanced techniques for table-text hybrid question answering (TextTableQA) are investigated by \citet{luo2023hrot_8}. Its main goal is to improve the way LLMs handle complex, numerical problems that come from combining text and table data. The problem of providing numerical answers over hybrid data consisting of tables and text (TextTableQA) is discussed in the study. Key research areas in this subject include the recent advent of LLMs and their In-Context Learning and Chain-of-Thought (CoT) prompting. The study presents Hybrid Prompt Strategy and Retrieval of Thought (HRoT) for TextTableQA, a novel prompting technique. By using this technique, the model's performance with hybrid data is greatly enhanced by encouraging the model to adopt retrieval thinking. In specifically for financial datasets, the research introduces a unique HRoT approach that greatly improves the performance of LLMs in responding to complex numerical inquiries in table-text hybrid situations. Combining retrieval thinking with a hybrid prompt technique, the method demonstrates enhanced efficacy while processing complex hybrid data.

\subsection{Accessibility and User-Friendly Approaches to TQA}

\textbf{TableQuery} focuses on making TQA more accessible and efficient for non-technical users, emphasizing ease of use and the ability to handle large datasets without requiring extensive training or technical expertise.

This paper's primary objective is to make it possible for users—particularly non-technical users to utilize natural language queries to evaluate big datasets in tabular form. This strategy aims to solve the shortcomings of current approaches that can't effectively handle enormous datasets or require a lot of training. TableQuery \citep{abraham2022tablequery_6} transforms natural language inquiries into structured queries by utilizing deep learning models that have been pre-trained for answering questions on free text. There's no need to fit the whole dataset into memory or serialize databases repeatedly when using these searches on databases or spreadsheets. TableQuery is a big step in improving the efficiency and accessibility of data querying, especially for non-technical users and huge datasets. Its performance is dependent on the capabilities of current question answering models, though, and a thorough evaluation of its ability to handle a wider range of questions and data types is still pending.

\section{Proposed Solution}
There are challenges when querying big databases with natural language. Large-scale, complicated repositories of data with a vast range of values and properties are characteristics of real-world systems. For instance, \textit{“What would be the average profit from selling OneView to Allianz in ANZ”?} from B2B Sales data \footnote{\url{https://www.kaggle.com/datasets/nnthanh101/aws-saas-sales}}. Under the assumption that a typical user wouldn't explicitly mention the precise column names, values, origins, the LLM in this scenario must determine that ANZ is a subregion, OneView is a product, and Allianz is a customer. One popular strategy, which ChatGPT ADA \footnote{\url{https://openai.com/index/improvements-to-data-analysis-in-chatgpt/}} also does, is to supply the top 5 rows of the dataset, from which the LLM builds its hypothesis and creates the DB query, to aid it in understanding the structure of the data. Even though big models like the gpt-4 can establish the correct associations when given descriptive and unambiguous column names, they frequently falter when faced with high cardinal characteristics, cryptic names, or columns that include lots of data. However, it is quite difficult for tiny and mid-sized models, such as the 7B and 13B parameters and their quantized equivalents, to decipher these assumptions and provide a logically and syntactically sound response.

Such complexity frequently makes it difficult for the LLMs to correctly deduce from a user's query what columns or values are intended, particularly when the query is written in an ambiguous or non-specific manner which is general in real In these situations, the models attempt to hide the wide possibilities by utilizing methods such as \texttt{contains()} or \texttt{LIKE} which leads to syntactically valid but logically erroneous queries. Similarly, if an attribute or value is spelled incorrectly, the model will generate a query based on the misspelled value and won't receive the intended answer because the query was logically incorrectly generated. Due to processing power restrictions and the intrinsic structure of these models, it is not feasible to transmit full schema metadata or whole datasets to the LLM, which exacerbates the limitations even more.

An innovative strategy that can bridge the gap between the rigid, structured world of database systems and LLMs' sensitive comprehension of natural language is needed to address these issues. To reduce guesswork and create database queries that contain exact values as present in the data, regardless of how the user mentioned them in the question, our framework introduces a solution through the implementation of Full-Text Search (FTS) on the metadata of the input table. By using the FTS, a Dynamic Schema of the table is created around attributes and values involved in the question to refine even loosely defined user questions.

Based on the same index that powers the FTS, we propose a Contextual Auto-complete function to further improve user's ability to explore and interact with big and complicated datasets. This helps the user craft precise queries as intended by providing real-time dropdown suggestions as they type the question. It also lessens the need to frequently consult the data to find precise attribute and/or value names, which saves time and friction when asking the right question. Additionally, it aids in solving the cold start issue that arises when a user is given a blank text box without much background information about the material. The proposed methodology for RoundTable is depicted above in Fig. 1


\section{Implementation}

RoundTable framework makes use of open-sourced Large Language Models (LLMs) from Hugging Face \footnote{\url{https://huggingface.co/}} and LM Studio \footnote{\url{https://lmstudio.ai/}} for managing model operations, which forms the base of our experimental setup. Evaluation of the framework is performed on datasets from various sectors like supply chain, physical properties, and sales to facilitate a comprehensive data analysis. 
Initially, the system processes the data by identifying and categorizing attributes, assigning data types, and isolating unique elements. This preprocessing helps in creating an efficient index for better data retrieval.
When users interact with the system, an autocomplete feature suggests possible attributes and values as they type their queries. The process begins with the extraction of key words from the user input. This leads to metadata pruning to refine the search. The system identifies relevant attributes directly and indirectly, employing advanced detection methods. This generates a customized prompt used by the LLM to produce the final database queries. This method ensures that the system is user-friendly and capable of handling complex queries effectively across different domains.

\begin{algorithm}
\footnotesize
\caption{Index Creation}
\begin{algorithmic}[1]
\State \textbf{Input:} User Table $T$
\State \textbf{Output:} Inverse Index

\Function{Create\_Index}{$T$}
    \State $A, D \gets \text{Get\_Attribute\_Names\_And\_Types}(T)$
    \State $C \gets \text{Filter\_Categorical\_Attributes}(A, D)$
    \For{each attribute $a$ in $C$}
        \State $M \gets \text{Extract\_Unique\_Values}(T[a])$
    \EndFor
    \State $S \gets \text{Generate\_Synonyms}(A)$
    \State $I \gets \text{Build\_Inverse\_Index}(A, D, S, \{M\})$
    \State \Return $I$
\EndFunction
\end{algorithmic}
\end{algorithm}

\subsection{Attribute Extraction}

The initial phase of our approach starts with the careful determination of the structure of the input table, T. This is an important phase that is necessary to understand the underlying schema and configuration of T. It lays the groundwork for understanding the subsequent stages of query processing and indexing. Specifically, we use a function called \texttt{GET\_ATTRIBUTE\_NAMES\_AND\_TYPES}, which is intended to iterate over T in a systematic manner. The main goal of this method is to create an exhaustive list of attribute names (shown as `A') and the corresponding data types (shown as `D'). Because it gives a clear grasp of T's schema, which is essential for any further data processing and analysis activities, this phase is very important. By doing this, we create a basic understanding of the structure and content of the table, laying the groundwork for effective data management and use in subsequent activities.

\subsection{Categorization of Attributes}

Once the attribute extraction phase is completed, we employ a particular filtering procedure that we designate as `C', which is meant to discern categorical properties. In this stage, the function \texttt{FILTER\_CATEGORICAL\_ATTRIBUTES} is essential. It carefully goes over attribute set `A', using a criterion based on `D' to extract attributes that fall within the `Categorical' heading. The classification of qualities as `Categorical' suggests that they are made up of distinct classes/categories, which makes them extremely useful for querying. This classification is especially important since categorical data makes information easier to organize and retrieve, which speeds up the analytical process.

\subsection{Value Indexing}

Determining the structure of categorical data is a significant step once categorical features have been identified and isolated. For every categorical column, the function \texttt{EXTRACT\_UNIQUE\_VALUES} is specifically engineered to do a thorough analysis and extraction of a single set of values, represented by the letter `M'. These distinct variables become essential building blocks that are directly connected to query search phrases while building the search index. To precisely target and retrieve data based on certain query criteria, it is imperative that these unique values be extracted and identified. This process not only enhances the efficiency of data querying but also significantly contributes to the robustness of the search index by ensuring that each query term is accurately mapped to distinct, relevant data points within the categorical dataset.

\subsection{Synonym Generation}

Our technique considers the wide range of queries made by users and makes creative use of a Large Language Model (LLM) by implementing the \texttt{GENERATE\_SYNONYMS} function. The purpose of this function is to increase the search power of the properties in our dataset. It functions by using each attribute name ('a') in the set 'A' to produce a carefully selected set of synonyms ('S') that are especially meant to make sense in a commercial setting. This improvement is purposefully made to strengthen the index's robustness so that it can efficiently handle a broad variety of linguistic phrases that are utilized in queries. The user experience and overall accuracy of search results are improved when business-relevant synonyms are added to the search index. This makes a major improvement to the system's capacity to comprehend and respond to the different language users may use.

\subsection{Index Construction}

An essential component in creating the inverse index, represented by the letter I, is the \texttt{BUILD\_INVERSE\_INDEX} function. This function combines several parts, including attribute names, data types, synonyms, and unique values. The fundamental framework that greatly improves our query processing capabilities is this integration. It serves as the framework for quick and precise data retrieval procedures. Furthermore, this index is essential for enabling the dynamic creation of schemas, which permits adaptability to changing data needs. This mechanism plays an important role in enhancing the efficacy and efficiency of data processing in our system.

\begin{algorithm}
\footnotesize
\caption{Dynamic Schema Generation and Query Formation}
\begin{algorithmic}[1]
\State \textbf{Input:} User Query $Q$, Inverse Index $I$, Static Prompt $P$
\State \textbf{Output:} Generated Database Query $DQ$

\Function{GENERATE\_DYNAMIC\_SCHEMA} {$Q$, $I$, $P$}
    \State $K \gets \text{EXTRACT\_KEYWORDS} (Q)$
    \State $D, Ind \gets \text{SEARCH\_INDEX}(K, I)$
    \State $D \gets \text{CREATE\_PROMPT}(D, Ind)$
    \State $DQ \gets \text{Formulate\_QUERY}(D, P)$
    \State \Return $DQ$
\EndFunction
\end{algorithmic}
\end{algorithm}

\subsection{Pre-Processing User Query}

Process action starts in the first stage when the \texttt{EXTRACT\_KEYWORDS} function is activated. This essential feature is crafted to handle the pre-processing of the user's query, query Q. Its main goal is to sort the query to locate and separate the essential elements, or important keywords K. The function uses a variety of sophisticated NLP approaches to do this. These advanced techniques play a critical role in honing the question and making sure it is set up for success in the next steps. This pre-processing stage is essential for improving keyword matching efficiency and laying the groundwork for precise information retrieval or outcomes based on the extracted keywords. This methodological approach emphasizes how important it is to identify keywords precisely to optimize the system's search and analysis operations.

\subsection{Keyword Matching and Schema Generation}

To find terms relevant to the user's search request, the \texttt{SEARCH\_INDEX} function carefully searches through an index as the first stage in query processing. This is an important stage because it establishes the groundwork for later processes by detecting both obvious and subtle characteristics that might be important to the question. After that, the \texttt{CREATE\_PROMPT} function makes use of the recognized attributes (direct, which are directly related to the query, and indirect, which are relevant through associative relationships). In this case, a dynamic schema (designated as D) is created. This schema is especially designed to adjust to the needs of every query; it is not static. This design guarantees that the data structure is both flexible and immediately relevant to the user's query by considering the varied qualities that were previously determined. This improves the accuracy and efficiency of the data retrieval process.

\subsection{Integrating with Static Prompt}

Using the \texttt{FORMULATE\_QUERY} function, a dynamic schema (designated as D) is integrated with a static prompt (specified as P) beforehand. The direction of a large language model's capabilities is greatly influenced by this integration. It guarantees that the model produces semantically valid and syntactically correct database queries, which adhere to the conventions of database query language structure. Semantic correctness is the ability of the query to understand the meaning and purpose of the user's request. Through the combination of the dynamic schema's flexibility and the static prompt's consistency, this method efficiently makes use of the large language model's comprehension and processing capability to generate queries that precisely represent the intended information retrieval objectives. This technique emphasizes how critical it is to have a meticulously defined input to properly utilize large language models in scenarios involving database queries.

\subsection{LLM-Based Query Formation}

A major use of LLMs in bridging the gap between human language and machine executable code is demonstrated by the \texttt{GENERATE\_DB\_QUERY} function. It leverages the superior capabilities of LLMs to read and convert natural language into a structured database query (DQ) by receiving input in the form of a dynamic schema together with a static prompt. Understanding the subtleties of the input, mapping it against the designated schema, and then creating a query that precisely reflects the original request in a syntax that can be immediately performed by a database management system comprise this translation process. This demonstrates not only how flexible LLMs are in handling and transforming unstructured data into structured form, but also how they may improve the usability and accessibility of database interactions for people who lack extensive technical knowledge of database query languages.

\section{Results}

\begin{table*}
\centering
\begin{tabular}{l ccc c c}
 & & \textbf{Difficulty Level} & \\
\toprule
\textbf{Dataset} & \textbf{Easy} & \textbf{Medium} & \textbf{Hard} & \textbf{Category} & \textbf{Total} \\
\midrule
\textbf{B2B Sales} & 50 & 50 & 50 & Generic & 150 \\
 & 60 & 60 & 60 & Value-Based & 180 \\ 
\midrule
\textbf{Physical Gem Properties} & 50 & 40 & 50 & Generic & 140 \\
 & 60 & 60 & 60 & Value-Based & 180 \\ 
 \midrule
\textbf{Supply Chain} & 50 & 50 & 50 & Generic & 150 \\
 & 70 & 80 & 50 & Value-Based & 200 \\ 
 \midrule
 \textbf{Super Market Sales} & 60 & 20 & 30 & Generic & 110 \\
 & 110 & 220 & 120 & Value-Based & 450 \\
 \bottomrule
\end{tabular}
\caption{\label{citation-guide}
Distribution of types of questions involved in the benchmark dataset. Generic and Value-Based questions distributed across Easy, Medium, and Hard.
}
\end{table*}

\subsection{Dataset}
In order to evaluate the performance of the framework we required a dataset of natural language quantitative questions that are representative of the real world business scenarios.
Most of the datasets currently available that we know of are small tables mostly extracted from web pages. In order to overcome these issues we have created a dataset of 1500 quantitative questions from a variety of domains including supply chain \footnote{\url{https://www.kaggle.com/datasets/shreyashjaiswalshrey/supply-chain-data-fashion-beauty-startup}}, B2B sales \footnote{\url{https://www.kaggle.com/datasets/nnthanh101/aws-saas-sales}}, B2C sales \footnote{\url{https://www.kaggle.com/datasets/akashdeepkuila/big-mart-sales}}  and physical properties \footnote{\url{https://www.kaggle.com/datasets/shivam2503/diamonds}}.
Our aim is to cover a broad set of scenarios covering various ways users could possibly interrogate the data to extract insights.
To achieve the same we divide the questions into 2 broad categories.
\begin{enumerate}
    \item \textbf{Difficulty Level}
    \begin{enumerate}
        \item Easy
        \item Medium
        \item Hard
    \end{enumerate}
    \item \textbf{Category}
    \begin{enumerate}
        \item Generic
        \item Value-Based
    \end{enumerate}
\end{enumerate}
\textit{Difficulty Level} divides question into 3 sub-categories of varying complexity of questions while \textit{Category} further divides them into 2 categories.

\begin{itemize}
    \item \textbf{Generic}: These are questions which might include any specific column name but not any specific values like \textit{“What is the size of the table?”} or \textit{“What is the total revenue?”}
    \item \textbf{Value-Based}: These questions contain specific values often not in full or properly spelt like \textit{“What was the total revenue from Dave of Costco?”}
\end{itemize}

Distribution of questions across \textit{Difficulty Level} and \textit{Category} for all the included tabular datasets are shown in the Table 1 below.

Questions are generated with the help of ChatGPT by providing samples of the tabular datasets in an iterative fashion to cover the full spectrum of the table. Generated questions are then carefully filtered by human experts to keep quality questions and remove inferior and duplicate ones. Some questions, especially Value-Based are modified to make them as close to how a business user would ask. For example a business user would not probably mention the exact column name or mention the exact value within quotes. To create such questions ChatGPT generated questions are modified to achieve the same.

In order to cover the various ways different users would ask a same question we generate 10 variations of an original question using textual data augmentation techniques like synonym replacement, word deletion, word position swapping, sentence shuffling and back translation.
This ensures that the evaluation is designed not only to assess it's code generation capabilities but also to assess it's ability to understand the user's ask from a natural everyday language query in plain English.

A sample of original questions as well as examples of variations have been included in the Appendix which would give an idea to the readers.
\subsection{Model}

All the evaluation on the above datasets have been done using the Mistral-7B-Instruct-v0.2 \footnote{\url{https://mistral.ai/news/announcing-mistral-7b/}} model’s 4-bit quantized version to make sure the evaluation is easily reproducible and accessible to a large audience because of small compute footprint of the model. Model was served using LM Studio \footnote{\url{https://lmstudio.ai/}} application.
Evaluation is manually done by human experts using the accompanied application.

\subsection{Evaluation}
The above-mentioned datasets have been evaluated at different levels and categories of Questions and the graph below depicts the performance of RoundTable.

\begin{figure}[htp]
    \centering
    \includegraphics[width=8cm]{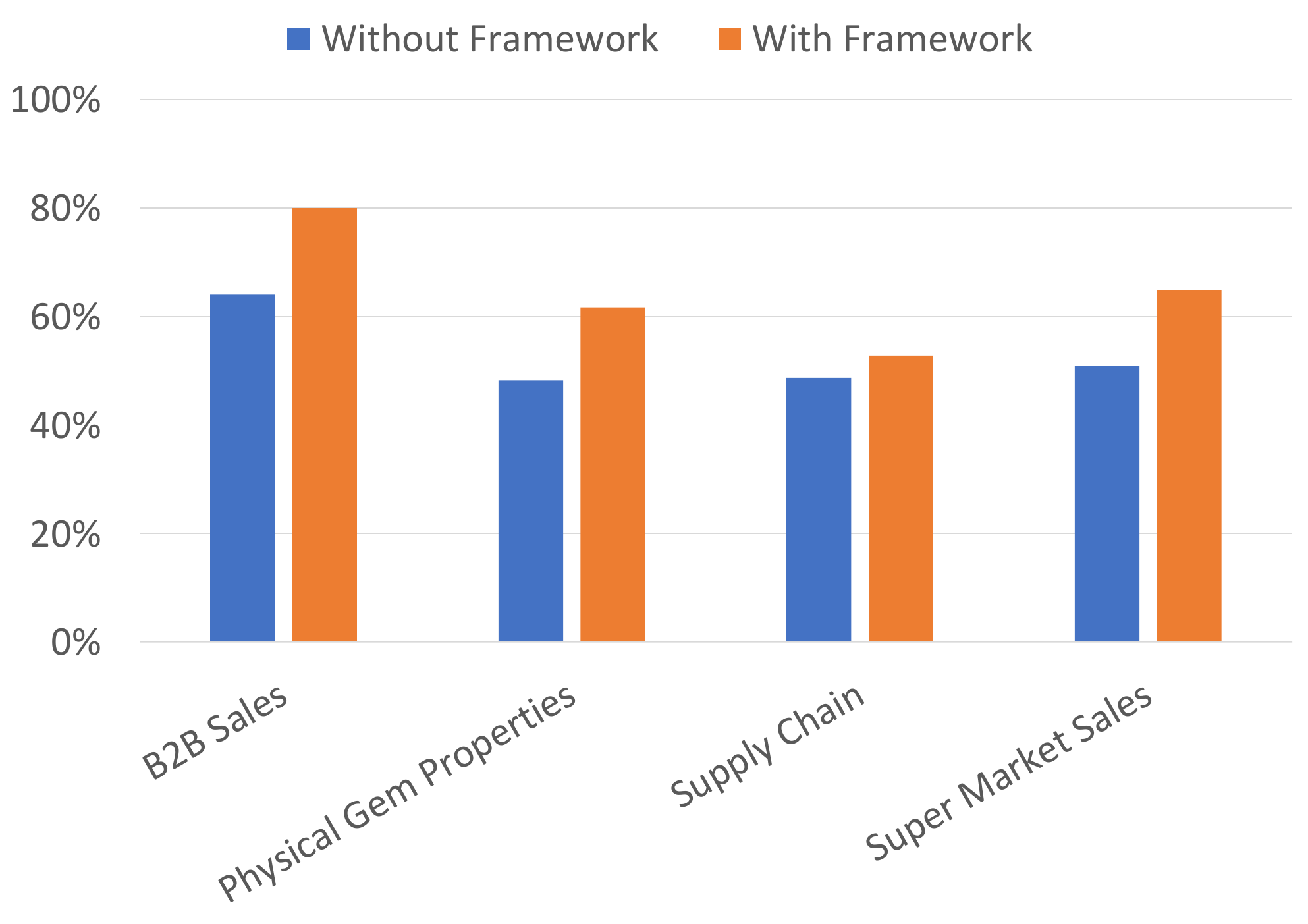}
    \caption{Performance of the model with the framework v/s without it on Generic questions. A slightly better performance can be observed when using the framework on questions which do not include specific data vocabulary.}
    \label{fig:fig2}
\end{figure}

\begin{figure}[htp]
    \centering
    \includegraphics[width=8cm]{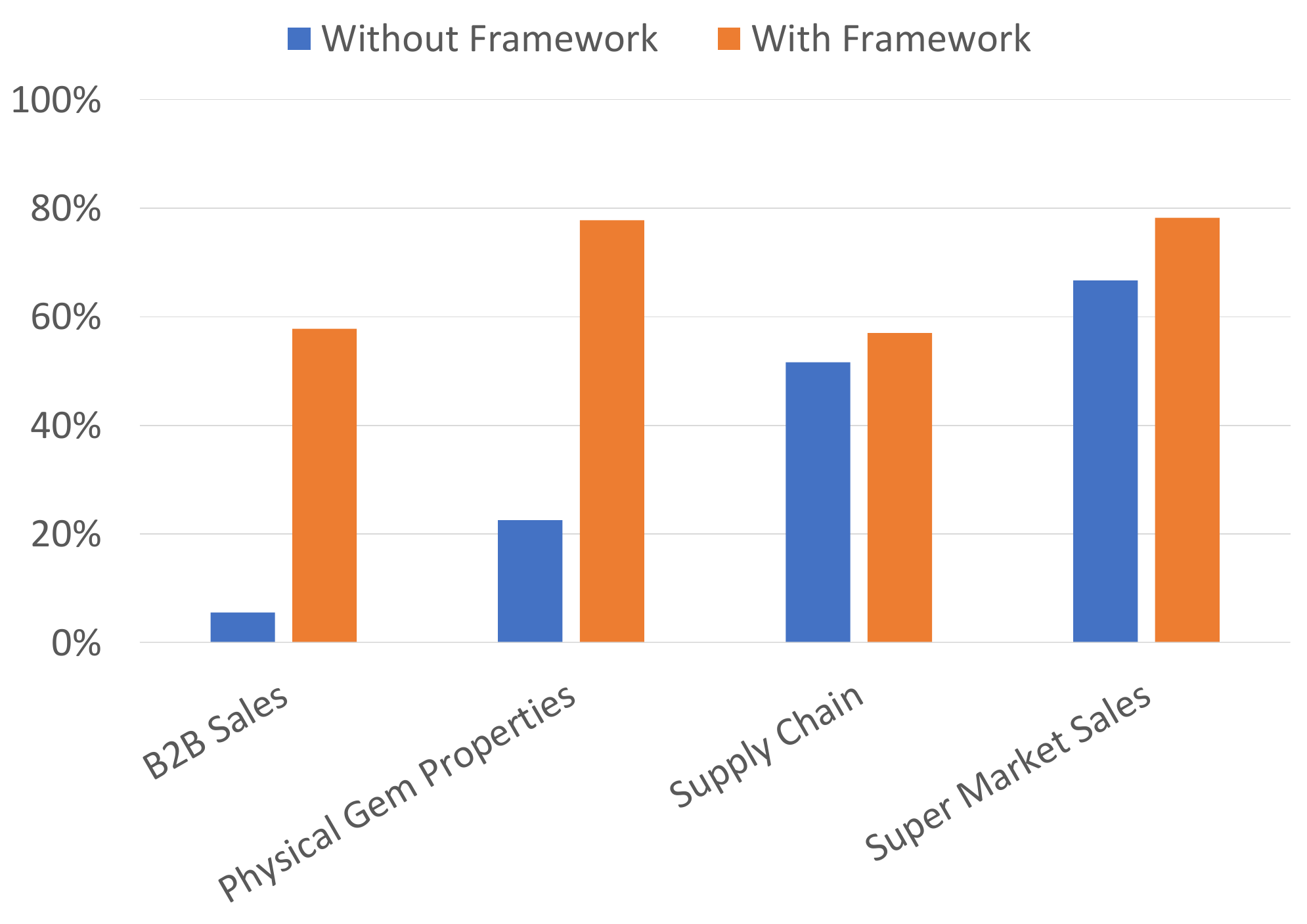}
    \caption{Performance of the model with the framework v/s without it on Value-Based questions. Significant improvement can be observed in questions where specific data vocabulary is involved.}
    \label{fig:fig2}
\end{figure}

Figure 2 presents a comparison of Generic questions across four datasets with and without RoundTable framework. Their is a slightly performance when the framework is used across datasets except for B2B Sales dataset.

Figure 3 presents a comparison of Value-Based questions across four datasets with and without RoundTable framework. In these questions their is a significant increase in accuracy observed when using the framework. This is  the area where the use of framework makes a significant difference compared to using vanilla LLMs.

Overall, the Framework's adoption has led to a substantial increase in accuracy across the board.

\begin{figure}[htp]
    \centering
    \includegraphics[width=8cm]{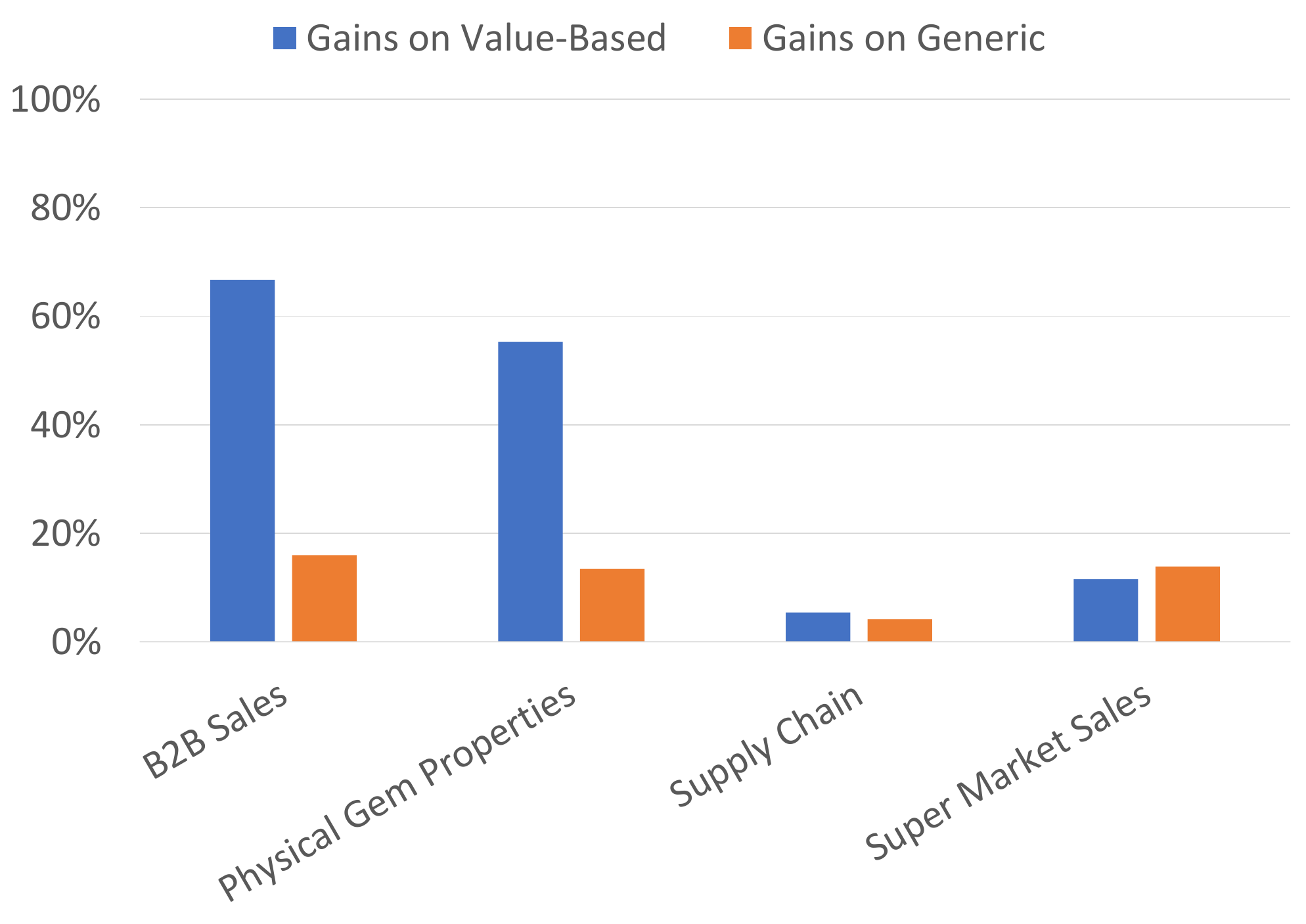}
    \caption{Overall gain in accuracy observed across datasets in both Generic and Value-Based questions when using the RoundTable framework. It is evident from the graph that higher gains are observed in Value-Based queries.}
    \label{fig:fig3}
\end{figure}

Figure 4 illustrates the overall gain achieved using the RoundTable approach versus without it. The B2B Sales dataset experienced the most significant increase, with a gain of 66.67\% on Value-Based questions, while the Supply Chain showed a smaller improvement, with gains of 5.40\% on Value-Based and 4.13\% when addressing Generic Questions. In scenarios involving Value-based Questions, the B2B Sales data showed 16\% gain, while the Physical Gem Properties recorded gains of 55.23\% on Value-Based questions.

\section{Conclusion}
In conclusion, the paper presents a transformative approach to Table Question Answering (TQA) by introducing a Full-Text Search (FTS)-enhanced dynamic schema and a Contextual autocomplete feature. These innovations address the current limitations in query precision and interaction with complex datasets. By narrowing the language model's search space and offering real-time query suggestions, the framework ensures a higher degree of query accuracy and user efficiency. The comprehensive evaluation across diverse datasets confirms the framework's effectiveness, showcasing significant improvements in query precision compared to traditional methods. This marks a considerable stride in making complex data querying accessible to users irrespective of their technical expertise, facilitating a more intuitive and efficient data interrogation process. The framework's ability to adapt to various domains and its scalability signals a promising direction for future research and applications in natural language database querying.

\bibliographystyle{acl_natbib}
\bibliography{custom}

\appendix
\onecolumn
\section{Appendix}

In this segment, extra information such as further findings, methodical explanations, and so on is offered to enhance the reader's comprehension of the ideas covered within the given study.

\subsection{RoundTable Dataset}
Here we present a sample of the dataset covering all possible combination of questions across three difficulty levels, that are Easy, Medium, and Hard as well as both types of questions, Generic and Value-based to give an idea of what kind of user queries have been used in evaluating the proposed work.

\addcontentsline{toc}{section}{Appendix}

\begin{table*}[htbp]
\centering
\begin{tabular}{P{8cm} P{2cm} P{2cm} P{2cm}}
    \toprule
    \textbf{Question} & \textbf{Difficulty Level} & \textbf{Dataset} & \textbf{Category} \\ 
    \midrule
    What is the highest single transaction value in terms of sales? & Easy & B2B Sales & Generic \\ 
    \hline
    How does the average profit per sale vary across different regions? & Medium & B2B Sales & Generic \\ 
    \hline
    Segment customers based on their total sales volume and determine the average profit for each segment. & Hard & B2B Sales & Generic \\ 
    \hline
    What is the total sales made by APJ-2023-127341? & Easy & B2B Sales & Value Based \\ 
    \hline
    In enterprise show the average sales for all healthcare customer & Medium & B2B Sales & Value Based \\ 
    \hline
    Analyze the average sales trend of the Data Smasher across different subregions & Hard & B2B Sales & Value Based \\ 
    \hline
    What is the average stock level across all SKUs? & Easy & Supply Chain & Generic \\ 
    \hline
    What is the average revenue generated per customer demographic? & Medium & Supply Chain & Generic \\ 
    \hline
    Analyze the variation in manufacturing costs by location. & Hard & Supply Chain & Generic \\ 
    \hline
    What percentage of products are categorized under skincare? & Easy & Supply Chain & Value Based \\ 
    \hline
    How does the manufacturing lead time vary between Supplier 1 and Supplier 3? & Medium & Supply Chain & Value Based \\ 
    \hline
    How does the choice of transportation mode affect the lead time for deliveries to Delhi? & Hard & Supply Chain & Value Based \\ 
    \bottomrule
\end{tabular}
\end{table*}

\onecolumn
\section{Data Augmentation}
Original Question: What is the standard deviation of the price for fair vvs2 diamonds with color F?
\\[1em]
Augmented Variations:
\begin{itemize}
    \item What is the standard deviation in price for vvs2 diamonds that are color 'F'?
    \item Can you provide the standard deviation of prices for vvs2 diamonds with the 'F' color grade?
    \item What is the price standard deviation for diamonds with 'F' color and vvs2 clarity?
    \item How much is the standard deviation of the prices for vvs2 clarity diamonds with 'F' color?
    \item Please calculate the price standard deviation of vvs2 diamonds with a color grade of 'F'.
    \item What's the standard deviation of the prices for diamonds with clarity vvs2 and color 'F'?
    \item How do the prices vary, in terms of standard deviation, for vvs2 diamonds that have an 'F' color?
    \item Could you tell me the standard deviation of the pricing for vvs2 grade diamonds with an 'F' color?
    \item What is the standard deviation of the 'F' colored vvs2 diamonds' price?
    \item Determine the standard deviation of prices for diamonds that are vvs2 clarity and 'F' in color.
\end{itemize}

\end{document}